\documentclass[letterpaper, preprint, paper,11pt]{AAS}	

\usepackage{bm}
\usepackage{amsmath}
\usepackage{amsfonts}
\usepackage{float}
\usepackage{subfigure}
\usepackage[colorlinks=true, pdfstartview=FitV, linkcolor=black, citecolor= black, urlcolor= black]{hyperref}
\usepackage{overcite}
\usepackage{footnpag}			      	
\usepackage{comment}

\PaperNumber{26-777}

\begin{document}

\title{Spacecraft Rendezvous Trajectory Generation with Modular Constraints via Diffusion Model Composition}

\author{Mariko A. Storey-Matsutani\thanks{Graduate Student, Department of Aeronautics and Astronautics, Massachusetts Inst. of Technology, Cambridge MA},  
Richard Linares\thanks{Associate Professor, Department of Aeronautics and Astronautics, Massachusetts Inst. of Technology, Cambridge MA}}

\maketitle{}

\begin{abstract}
Emerging mission classes such as on-orbit servicing, satellite inspection, and active debris removal require trajectory design methods that are adaptable to a variety of mission scenarios. We present a diffusion-based trajectory generation approach for rendezvous and proximity operations (RPO) that enables flexible configuration of mission constraints. First, individual energy-based diffusion models are trained to satisfy distinct constraints such as approach cone and sensor line-of-sight from a set of optimized trajectories. Then, at inference time, the learned energy models can be composed with one another, or with an analytically defined energy field, to enforce specific constraint combinations. We validate this framework with the composition of a learned approach cone model and a learned sensor line-of-sight model, as well as a learned approach cone model and synthetic obstacle avoidance model, both of which yield constraint satisfaction rates that are within 1 percentage point of the single-constraint models or higher. These results indicate that our compositional diffusion framework can provide a modular approach to RPO trajectory design and enable reconfiguration for new constraint combinations without requiring model retraining.

\end{abstract}

\section{Introduction}
In recent years, rapidly growing interest in space-based infrastructure has accelerated the need for advanced trajectory design methods. In particular, emerging mission classes such as on-orbit servicing, active debris removal, and in-space assembly have emphasized a focus on rendezvous and proximity operations (RPO) maneuvers, which require spacecraft to operate in close proximity while satisfying dynamical and safety constraints \cite{starek_fast_2017}. The ongoing NASA Artemis program and commercial space station development efforts further motivate the need for robust autonomous RPO capabilities to expand human presence in space, as tasks like docking and resupply are prime examples of critical RPO maneuvers \cite{smith_artemis_2020}.

The relative motion dynamics for RPO regimes are commonly modeled using linearized formulations such as the Clohessy–Wiltshire (CW) or Hill equations \cite{starek_fast_2017, clohessy1960terminal}. However, practical missions require further optimization methods to yield fuel-optimal trajectories that are typically prioritized in space missions \cite{miele_optimal_2007}. Despite the benefit of these linearized models, trajectory design for RPO is fundamentally challenging because the close coordination of spacecraft introduces tightly coupled constraints that require full, six degree-of-freedom motion planning. The consideration of approach cone, sensor line-of-sight (LOS), plume impingement avoidance, and collision avoidance constraints can introduce nonconvex feasible regions in the state-control space \cite{starek_fast_2017}.

As shown in classical optimal control literature, one approach to solving such nonconvex trajectory optimization problems is sequential convex programming (SCP), which has been applied to RPO mission applications.\cite{malyuta_fast_nodate} SCP works by iteratively approximating a convex subproblem of the original nonconvex problem formulation, which enables the application of traditional convex optimization methods \cite{mao_successive_2016,szmuk_successive_2017}. While SCP has demonstrated strong performance in many path planning scenarios, its convergence can be sensitive to initialization and the addition or modification of constraint sets requires reformulation of the convexification strategy. \cite{kim_guided_2022} As missions increasingly rely on coordinated, multi-spacecraft operations in cislunar and Earth-orbit environments, there is a need for modular, flexible tools for trajectory design that can adapt to changing mission needs.

Given these limitations, recent work has explored learning-based alternatives for the generation of spacecraft trajectories. Reinforcement learning \cite{chang_reinforcement_nodate}, imitation learning \cite{shao_imitation_2026}, and transformer-based architectures \cite{briden_improving_nodate, guffanti_transformers_2024, takubo_towards_2025, jain_multi-phase_2025} have been used to approximate optimal policies or directly generate feasible trajectories, as well as to warm-start traditional solvers for nonconvex optimal spacecraft control. However, many existing approaches exhibit limited generalization across varying constraint configurations. As a result, changes in constraint geometry or combinations typically necessitate retraining, thus reducing flexibility and limiting deployment in adaptive mission contexts \cite{briden_diffusion_2025,li_diffusolve_2024}.

To this end, existing research has begun to explore the application of diffusion models as an alternative approach to generative trajectory design \cite{briden_diffusion_2025, presser_diffusion_2024, li_diffusolve_2024}. Diffusion probabilistic models are a type of generative model capable of representing complex multimodal distributions through an iterative denoising process \cite{ho_denoising_2020,song_score-based_2021}. Originally developed for high-dimensional data such as images\cite{ho_denoising_2020}, diffusion models learn probabilistic distributions over solution spaces and allow conditioning mechanisms that are well-suited for trajectory generation problems \cite{janner_planning_2022}. In particular, the formulation of diffusion models as energy-based models further supports model composition and posterior conditional sampling, enabling constraint-flexible trajectory generation without model retraining \cite{du_reduce_2024}.

Diffusion-based models have recently been applied to several spacecraft trajectory problems, including powered descent guidance, low-thrust transfers, and ballistic transfer generation \cite{briden_diffusion_2025,briden_compositional_2025,graebner_global_2025,presser_diffusion_2024}. These works show that diffusion models can learn complex solution manifolds, provide diverse and high-quality initial guesses, and accelerate warm-starting of indirect and direct optimal control solvers \cite{briden_diffusion_2025,li_diffusolve_2024,graebner_global_2025}. However, most existing applications focus on single-vehicle scenarios with relatively structured constraints, and do not directly address RPO-specific constraints like approach cones, collision avoidance, and sensor line of sight. 

This gap motivates the compositional diffusion framework for RPO trajectory generation developed in this work. Instead of training a single complex model to satisfy all mission constraints simultaneously, we demonstrate the decomposition of the constraint space into modular components and train individual diffusion models on distinct constraint classes (e.g. approach cone, sensor line-of-sight). At inference time, these models are composed to generate trajectories that satisfy multiple constraints concurrently, enabling reconfiguration to new RPO mission constraint profiles while reducing retraining requirements.

\section{Methodology}

We propose a two-phase framework for generating warm-start trajectories for rendezvous and proximity operations (RPO) under modular constraint profiles. In the first phase, individual diffusion models are trained on constraint-specific expert trajectories that are generated offline using sequential convex programming (SCP) via the SCPToolbox framework \cite{SCPToolboxCSM2022}. In the second phase, these diffusion models are composed and conditioned at inference time to generate trajectories that satisfy flexible constraint combinations. The following sections detail the key elements of this framework, including data generation, diffusion model training, and composition operations. 

\subsection{SCP-Based Data Generation}
Expert trajectories are generated using the Apollo CSM rendezvous example provided in SCPToolbox. \cite{SCPToolboxCSM2022} The relative translational dynamics between the two spacecraft are modeled using the linearized Clohessy--Wiltshire (CW) equations in the local orbital frame, while the rotational motion is represented through rigid-body attitude dynamics under thruster actuation \cite{SCPToolboxCSM2022}. The trajectory's initial state is randomized within a defined range to generate diverse rendezvous scenarios. 

The system state and control vectors are defined as

\begin{equation}
    x_k =
\begin{bmatrix}
r_k \\ v_k \\ q_k \\ \omega_k
\end{bmatrix}
\in \mathbb{R}^{13},
\qquad
u_k =
\begin{bmatrix}
f_k \\ f_k^{\mathrm{ref}} \\ f_k^{\mathrm{eq}}
\end{bmatrix}
\in \mathbb{R}^{33}.
\end{equation}

Here $r_k$ and $v_k$ denote the relative position and velocity in the local orbital frame, $q_k$ represents the spacecraft attitude quaternion using scalar-last convention, and $\omega_k$ denotes the angular velocity. The control vector contains the commanded thruster forces $f_k \in \mathbb{R}^{16}$, the reference thruster forces $f_k^{\mathrm{ref}} \in \mathbb{R}^{16}$, and a scalar slack variable $f_k^{\mathrm{eq}} \in \mathbb{R}$ used to enforce the minimum-impulse firing logic \cite{SCPToolboxCSM2022}. Expert trajectories are generated by solving the following trajectory optimization problem based on the Apollo CSM rendezvous scenario implemented in SCPToolbox \cite{SCPToolboxCSM2022}:

\begin{equation}
\begin{array}{r l}
\displaystyle
\min_{\{x_k,u_k\}_{k=0}^{N-1}, t_f, \Delta x_f}&
\sum_{k=0}^{N-1}
w_k \left(
\frac{\mathbf{1}^\top f_k}{f_{\max}}
+ \gamma_c \frac{f_k^{\mathrm{eq}}}{f_{\min}}
\right)\\
&\text{s.t.}  \\
\text{Dynamics:} &
x_{k+1} = f_d(x_k,u_k,t_f), \qquad k = 0,\ldots,N-2 \\

\text{Initial state:} &
x_0 = x_{\mathrm{init}} \\

\text{Docking condition:} &
x_{N-1} + \Delta x_f = x_{\mathrm{dock}} \\

\text{Terminal position tolerance:} &
\|\Delta r_f\|_\infty \le r_{\mathrm{tol}},
\qquad \xi^\top \Delta r_f = 0 \\

\text{Terminal velocity tolerance:} &
\|\Delta v_f\|_\infty \le v_{\mathrm{tol}} \\

\text{Terminal angular rate tolerance:} &
\|\Delta \omega_f\|_\infty \le \omega_{\mathrm{tol}} \\

\text{Attitude alignment:} &
q_{N-1}^\top q_f
\ge \cos\!\left(\frac{\theta_{\mathrm{tol}}}{2}\right) \\

\text{Thruster limits:} &
0 \le f_{k,i} \le f_{\max},
\qquad \forall i = 1,\ldots,16 \\

\text{Reference thrust limits:} &
0 \le f_{k,i}^{\mathrm{ref}} \le f_{\max},
\qquad \forall i = 1,\ldots,16 \\

\text{Control allocation slack:} &
\|f_k - f_k^{\mathrm{ref}}\|_1 \le f_k^{\mathrm{eq}},
\qquad k = 0,\ldots,N-1 \\

\text{Final time bounds:} &
t_f^{\min} \le t_f \le t_f^{\max} \\

\text{Minimum thruster firing constraint:} &
f_{k,i}
- \mathrm{OR}_\kappa\!\left(f_{k,i}^{\mathrm{ref}} - f_{\min}\right) f_{k,i}^{\mathrm{ref}}
\le 0 \\

\text{Reference thrust consistency:} &
\mathrm{OR}_\kappa\!\left(f_{k,i}^{\mathrm{ref}} - f_{\min}\right) f_{k,i}^{\mathrm{ref}}
- f_{k,i}
\le 0 \\

\text{Additional path constraints:} &
s(x_k,u_k;\kappa) \le 0
\end{array}
\end{equation}

We focus on defining the distinct constraints that were used to demonstrate the compositional nature of the proposed diffusion-based trajectory generation framework: the approach cone constraint and the sensor line-of-sight (LOS) constraint. Separate datasets are generated for each constraint, with a common set of base RPO constraints (dynamics, docking, thruster, and minimum-impulse logic) and the distinct additional path constraint $s(\cdot)$.

The approach cone constraint restricts the vehicle to approach the target from a specified direction once within a distance threshold $r_A$:

\begin{equation}
    \cos \theta_A
- \frac{r_k^\top \xi}{\|r_k\|}
\le 0, \qquad \text{when }  \, \|r_k\| < r_A.
\label{eq. appch}
\end{equation}

This constraint ensures that the spacecraft's relative position $r_k$ lies within a cone of half-angle $\theta_A$ aligned with the docking axis $\xi$ when the vehicle enters the terminal approach region of radius $r_A$. The distance-dependent activation is implemented through $\mathrm{OR}_\kappa$, a homotopy-parameterized approximation to the logical disjunction that enforces the angular restriction once the spacecraft is within the approach distance threshold \cite{SCPToolboxCSM2022}.

The sensor line-of-sight constraint seeks to maintain the chaser's target visibility over the full trajectory. This requires the body-fixed sensor axis to remain pointed toward the target within a sensor viewing angle, as defined by

\begin{equation}
\cos\theta_S - b \cdot \hat{r} \le 0 .
\label{eq. los}
\end{equation}

This constraint enforces that the angle between the chaser's body-fixed sensor axis $b$ and the target direction $\hat{r}$ remains below a threshold $\theta_S$ throughout the trajectory. 

The resulting trajectories are filtered for numerical validity and dynamic feasibility. To reduce the dimensionality of model training, the $16$ commanded RCS thruster forces are aggregated into a net three-axis force and torque vector, yielding a 6-dimensional control representation. Combined with the 13-dimensional spacecraft state, this yields a 19-dimensional feature vector discretized over 50 time steps. This feature vector is normalized per-feature to have zero mean and unit standard deviation ($\mu = 0, \sigma = 1$) prior to training. The parameters for the generated data are given in Table \ref{tab:dataset_summary}.

\begin{table}[H]
    \centering
    \caption{Dataset and Constraint Parameters}
    \label{tab:dataset_summary}
    \begin{tabular}{ll}
    \hline
    Parameter & Value \\
    \hline
    Initial position ranges (m) & ([45, 135], [-75, 75], [-75, 75])\\ 
    Final position (m) & (4.48, 0.05, 0.17) \\
    Time steps per trajectory & 50 \\
    Time of flight range (s) & [100, 1000] \\
    Approach radius $r_A$ (m) & 30 \\
    Approach cone half-angle $\theta_A$ & $10^\circ$ \\
    Sensor FOV half-angle $\theta_S$ & $50^\circ$ \\
    Approach Cone Dataset Size & 14259\\
    Sensor Line-of-Sight Dataset Size & 15786\\
    \hline
    \end{tabular}
\end{table}

\subsection{Diffusion-Based Trajectory Modeling}
Diffusion model training involves a forward ``noising" process that gradually corrupts the input data as a neural network learns the reverse ``denoising" process to reconstruct the input. While standard diffusion models apply independently sampled Gaussian noise to each state, this assumption neglects the strong temporal dependencies inherent in trajectory data. To address this limitation, we adopt the Dynamical Diffusion formulation introduced by Guo et al. \cite{guo_dynamical_2025}, which incorporates an explicit temporal coupling mechanism into the forward process.

Given a trajectory $x_0 = \{x_0^s\}_{s=0}^{S-1}$, the forward process constructs noisy samples by combining Gaussian diffusion with a recursive dynamics operator. At diffusion step $t$, the corrupted trajectory is defined as
\begin{equation}
x_{t}^{s}
=
\sqrt{\bar{\gamma}_t}
\left(
\sqrt{\bar{\alpha}_t}\,x_{s}^{0}
+
\sqrt{1-\bar{\alpha}_t}\,\epsilon_{t}^{s}
\right)
+
\sqrt{1-\bar{\gamma}_t}\,
x_{t}^{s-1},
\qquad
\epsilon_{t}^{s}\sim\mathcal{N}(0,I),
\end{equation}
where $\bar{\alpha}_t$ is the cumulative diffusion noise schedule and $\bar{\gamma}_t$ governs the strength of temporal coupling between adjacent trajectory states. As $\bar{\gamma}_t$ increases, the process places greater emphasis on local state information, while smaller values promote stronger temporal smoothing \cite{guo_dynamical_2025}.

This construction can be equivalently expressed through a dynamics transformation operator $\Phi(\cdot, \bar{\gamma}_t)$ applied recursively along the trajectory:
\begin{equation}
x^{\mathrm{dyn}} = \Phi(x, \bar{\gamma}_t), \qquad
x^{s,\mathrm{dyn}} =
\sqrt{1 - \bar{\gamma}_t}\,x^{s-1,\mathrm{dyn}} +
\sqrt{\bar{\gamma}_t}\,x^s,
\end{equation}
with initialization $x^{0,\mathrm{dyn}} = x^0$. The same operator is applied to both the clean trajectory and the Gaussian noise, yielding $x_0^{\mathrm{dyn}} = \Phi(x_0, \bar{\gamma}_t)$ and $\epsilon_t^{\mathrm{dyn}} = \Phi(\epsilon_t, \bar{\gamma}_t)$. This ensures that both signal and noise share consistent temporal structure \cite{guo_dynamical_2025}.

We adopt an energy-based formulation of diffusion models~\cite{du_reduce_2024}, in which the model learns a scalar energy function $E_\theta(x_t^{\mathrm{dyn}}, t)$ over the dynamics-transformed trajectory. The model is trained via standard denoising score matching, such that the scaled energy gradient approximates the temporally-conditioned noise via
\begin{equation}
\mathcal{L}(\theta)
=
\mathbb{E}_{x_0, t, \epsilon_t}
\left[
\left\|
\epsilon_t^{\mathrm{dyn}}
-
\frac{1}{\sqrt{1-\bar{\alpha}_t}}
\nabla_{x_t^{\mathrm{dyn}}} E_\theta(x_t^{\mathrm{dyn}}, t)
\right\|^2
\right].
\label{eq:loss}
\end{equation}

During sampling, we reverse the diffusion process iteratively from Gaussian noise. At each step $t$, the current sample is transformed via $x_t^{\mathrm{dyn}} = \Phi(x_t, \bar{\gamma}_t)$, and a noise estimate is obtained from the energy gradient:
\begin{equation}
\hat{\epsilon}_\theta^{\mathrm{dyn}}(x_t, t) = \frac{1}{\sqrt{1-\bar{\alpha}_t}}\nabla_{x_t^{\mathrm{dyn}}} E_\theta(x_t^{\mathrm{dyn}}, t).
\end{equation}
This is mapped back to untransformed coordinates via the inverse dynamics operator:
\[
\hat{\epsilon} = \Phi^{-1}(\hat{\epsilon}_\theta^{\mathrm{dyn}}, \bar{\gamma}_t), \quad \text{where} \quad x^s = \frac{x^{s,\mathrm{dyn}} - \sqrt{1-\bar{\gamma}_t}\,x^{s-1,\mathrm{dyn}}}{\sqrt{\bar{\gamma}_t}} .
\]
The clean trajectory prediction is then given by
\begin{equation}
\hat{x}_0 = \frac{x_t - \sqrt{1-\bar{\alpha}_t}\,\hat{\epsilon}}{\sqrt{\bar{\alpha}_t}},
\end{equation}
which is fed into a standard reverse diffusion update to obtain $x_{t-1}$. To accelerate sampling, we employ a DDIM (Denoising Diffusion Implicit Models) sampler \cite{song_denoising_2022}, which uses deterministic reverse steps instead of the stochastic updates of DDPM (Denoising Diffusion Probabilistic Models) \cite{ho_denoising_2020}. This enables fast sampling with 50-100 denoising steps while maintaining comparable sample quality.

The energy network $E_\theta$ is realized with a U-Net architecture, a standard choice for diffusion models due to its ability to capture multi-scale feature interactions \cite{ho_denoising_2020,dhariwal_diffusion_2021,ronneberger2015unet}. Trajectories are encoded as single-channel $50 \times 19$ tensors (50 time steps, 19 features) and processed through a symmetric encoder--decoder with convolutional blocks, layer normalization, and swish activations. Skip connections between matching encoder and decoder levels preserve fine temporal detail. The diffusion timestep $t$ is encoded via sinusoidal embeddings and additively injected into each block to condition the network on the noise level. The final convolutional layer produces a feature map that is spatially summed to yield the scalar energy $E_\theta(x_t^{\mathrm{dyn}}, t)$. The input gradient of this energy function ($\nabla_{x_t^{\mathrm{dyn}}} E_\theta$) serves as the score estimate used during training and sampling. Architectural and training hyperparameters are provided in Table~\ref{tab:unet_params}.

\begin{table}[H]
    \centering
    \caption{U-Net Architecture and Training Hyperparameters}
    \label{tab:unet_params}
    \begin{tabular}{ll}
    \hline
    Parameter & Value \\
    \hline
    \multicolumn{2}{l}{\textit{Architecture}} \\
    Input shape (time $\times$ features $\times$ channels) & $50 \times 19 \times 1$ \\
    Resolution levels & 4 \\
    Base channels & 32 \\
    Channel multipliers & $(1, 2, 4, 8)$ \\
    Block structure & $2 \times$ ($3 \times 3$ Convolution,\\
    & \quad \quad Layer Normalization,\\
    & \quad \quad Swish activation) \\
    Downsampling & $2 \times 2$ max pooling \\
    Upsampling & $4 \times 4$ transposed convolution \\
     & \quad \quad Stride 2 \\
    \hline
    \multicolumn{2}{l}{\textit{Training}} \\
    Diffusion steps & 1000 \\
    Noise schedule & Linear, $\beta \in [10^{-4}, 2\times10^{-2}]$ \\
    Temporal coupling $\bar{\gamma}_t$ & 0.5 \\
    Optimizer & Adam \\
    Learning rate & $1 \times 10^{-4}$ \\
    Batch size & 32 \\
    Training epochs & 100 \\
    EMA decay & 0.995 \\
    \hline
    \end{tabular}
\end{table}

\subsection{Constraint Composition and Conditioning}
The energy-based formulation enables inference-time composition of independently trained models without retraining \cite{du_reduce_2024}. In particular, multiple constraint models can be combined by manipulating their probability densities or corresponding energy functions during sampling. Two composition operators are considered in this work: product composition, which enforces simultaneous satisfaction of multiple constraints; and negation composition, which suppresses trajectory regions associated with undesirable constraint energies. Following Briden et al. \cite{briden_compositional_2025, briden_diffusion_2025}, we define product and negation composition as weighted sums of learned energy functions applied during DDIM sampling. Let $p_i(x)$ denote the trajectory distribution learned for constraint set $i$, with corresponding energy function $E_i(x)$. These operators are defined as
\begin{equation}
\text{Product:}\quad
 E_{\mathrm{prod}}(x_t^{\mathrm{dyn}}, t)
= \sum_{i=1}^{K} w_i E_i(x_t^{\mathrm{dyn}}, t),
\qquad
w_i > 0 
\end{equation}
\begin{equation}
\text{Negation:}\quad
 E_{\mathrm{neg}}(x_t^{\mathrm{dyn}}, t)
= \alpha E_1(x_t^{\mathrm{dyn}}, t)
- \beta E_2(x_t^{\mathrm{dyn}}, t),
\qquad
\alpha,\beta > 0
\end{equation}

where $K$ is the number of component models, $w_i$ are nonnegative composition weights, and $\alpha,\beta$ control the relative influence of attractive and repulsive energy terms \cite{du_reduce_2024, du_compositional_2020, briden_compositional_2025, briden_diffusion_2025}. 

In addition to the composition of learned energy models, we investigate the composition of a learned energy model with an analytically-defined energy field, which offers a simpler approach for constraints with predictable energy landscapes. For example, to encourage obstacle-avoidance behavior at inference time, we artificially generate an energy field whose values are highly negative at the obstacle location and decay in magnitude with radial distance. The synthetic obstacle energy is defined as

\begin{equation}
E_{\mathrm{obs}}(x)
= g \sum_{t}\left[-\operatorname{softplus}\left(-\frac{\|p_{\mathrm{phys}}(t) - c\|_2 - (R-\sigma)}{\sigma}\right)\right],
\label{eq:obs-energy}
\end{equation}

where $p_{\mathrm{phys}}(t)$ is the position of the spacecraft at time $t$ in physical units, $c$ is the obstacle center, $R$ is the obstacle radius, $\sigma$ is the softening length, and $g$ is the gain. All other dimensions besides position are given an energy value of zero. This synthetic energy field is then composed with a learned energy model via negation composition to yield a composed model that respects both the obstacle avoidance and learned model constraints.

Aside from full model composition, inference-time conditioning can be applied through trajectory inpainting. During the iterative denoising process, inpainting fixes selected feature values to prescribed targets at each step while the remaining features are updated by the learned reverse process. At each reverse step, the clean-trajectory estimate $\hat{x}_0$ is computed as above; the entries in the inpaint set $\mathcal{M}$ are overwritten with their target values, and the modified estimate is used in the subsequent reverse diffusion update, as in Equation \ref{eq:inpaint}. By repeatedly re-imposing the fixed values throughout sampling, the diffusion process generates trajectories that remain consistent with the specified constraints \cite{yang_planning_2023, li_aligning_2025}. We employ  this inpainting mechanism to  condition the final position of the trajectory to the docking site, while allowing the remaining portions of the trajectory to be generated by the diffusion model. The general inpainting formulation is

\begin{equation}
x_{t-1}^s =
\sqrt{\bar{\alpha}_{t-1}}\;\tilde{x}_0^s
+ \sqrt{1-\bar{\alpha}_{t-1}}\;\hat{\epsilon}^s,
\qquad
\tilde{x}_0^s =
\begin{cases}
y^s & \text{if } s \in \mathcal{M}\ \text{(inpaint region)} \\[0.4em]
\hat{x}_0^s & \text{if } s \notin \mathcal{M}\ \text{(non-inpaint region)}
\end{cases}
\label{eq:inpaint}
\end{equation}

\noindent
where $\mathcal{M}$ indexes the fixed inpaint region (e.g. the position at the final trajectory timestep) and $y^s$ is the overwriting trajectory data (e.g. the target position) \cite{lugmayr_repaint_2022}.

Together, these operations enable flexible inference-time manipulation of learned trajectory distributions, allowing constraint composition and partial trajectory conditioning without retraining the underlying diffusion models.

\section{Results}
This section evaluates the compositional diffusion-based framework for RPO trajectory generation. Our objective is to determine whether energy-based model composition and synthetic energy shaping can enforce combined constraints at inference time without any additional training. In addition to constraint satisfaction,  the models are evaluated on their ability to learn or preserve feasible trajectory dynamics and fuel optimality. Unless otherwise noted, all diffusion trajectories are generated with 50 DDIM steps and metrics are computed over 100 generated samples per model with a 50-step discretization per trajectory sample. 

\subsection{Individual Model Performance}
We first characterize each single-constraint model in isolation to establish that neither trivially satisfies the other's constraint, thereby motivating a meaningful composition study.  One model is trained on the approach cone constraint of Equation~\ref{eq. appch} and one on the sensor line-of-sight (LOS) constraint of Equation~\ref{eq. los}.

\begin{figure}[H]
    \centering
    \includegraphics[width=0.8\linewidth]{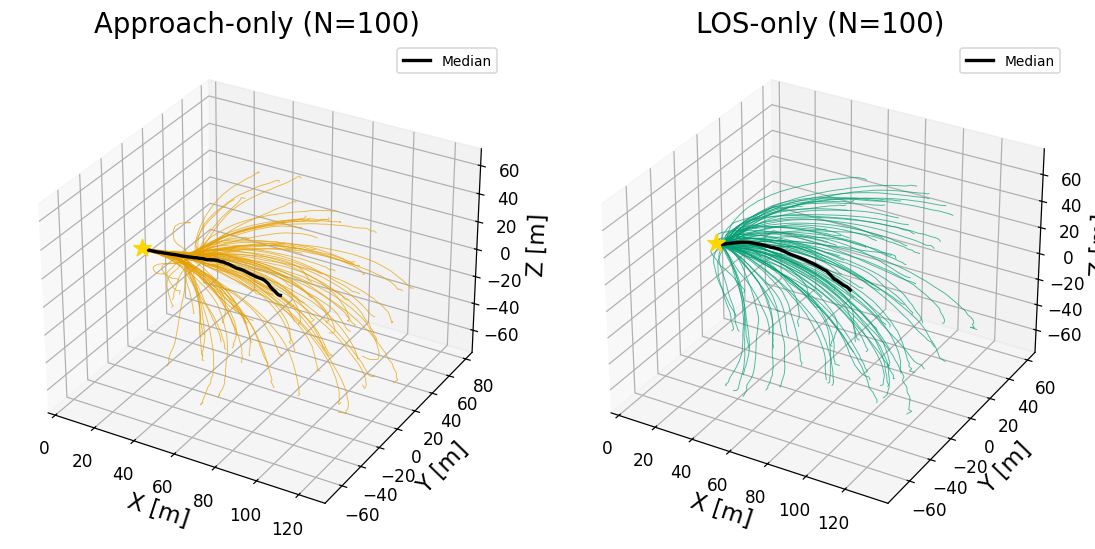}
    \caption{3D plot of Approach Cone-Only and LOS-Only Trajectory Samples}
    \label{fig:appchLOS_3d}
\end{figure}

The 100 samples drawn from each model are shown in Figure~\ref{fig:appchLOS_3d}, and per-constraint metrics are reported in Table~\ref{tab:constraint_errors} along with training data metrics for reference. The satisfaction metric measures the mean fraction of constraint-active timesteps satisfying the constraint, while the peak violation metric measures the mean (across samples) of the maximum angular constraint violation along each trajectory's constraint-active region. Each model closely satisfies the constraint on which it was trained, while exhibiting significant violations for the other. The approach cone model holds the approach cone over $96.1\%$ of constraint-active timesteps, while its sensor pointing is effectively unconstrained. The approach cone model satisfies the LOS constraint at only $24.2\%$ of constraint-active timesteps and exhibits a mean peak violation of $104.8^\circ$. Conversely, the LOS model satisfies the LOS constraint at $99.6\%$ of constraint-active timesteps but satisfies the approach cone at only $36.9\%$ of constraint-active timesteps with a mean peak violation of $20.2^\circ$. This confirms a compelling scenario for composition, since the two constraints are non-trivial and neither single model satisfies both constraints.

\begin{table}[H]
    \centering
    \caption{Constraint Satisfaction and Peak Violation Metrics for the Constraint-Specific Models and Training Data (N=100) }
    \label{tab:constraint_errors}
    \begin{tabular}{lcccc}
    \hline
     & \multicolumn{2}{c}{\textbf{Approach Cone Constraint}} & \multicolumn{2}{c}{\textbf{Sensor LOS Constraint}} \\
     & Satisfaction (\%) & Peak Violation ($^\circ$) & Satisfaction (\%) & Peak Violation ($^\circ$) \\
    \hline
    
    \textbf{Learned Models} & & & & \\
    Approach Cone & $96.1 \pm 9.6$  & $0.29 \pm 0.76$   & $24.2 \pm 7.7$ & $104.77 \pm 14.87$ \\
    Sensor LOS    & $36.9 \pm 27.5$ & $20.21 \pm 14.66$ & $99.6 \pm 3.0$ & $0.12 \pm 0.95$ \\
    \hline
    
    \textbf{Training Data} & & & & \\
    Approach Cone & $97.0 \pm 4.5$  & $0.32 \pm 6.69$   & $28.2 \pm 8.7$  & $104.58 \pm 13.89$ \\
    Sensor LOS    & $36.3 \pm 25.5$ & $20.70 \pm 14.49$ & $100.0 \pm 0.2$ & $0.02 \pm 1.45$ \\
    \hline
    \end{tabular}
\end{table}

In addition to constraint satisfaction, we investigate the dynamic feasibility of the generated samples. Since the denoising process of diffusion sampling simultaneously generates the full trajectory output, it is necessary to validate that the generated state and control values are self-consistent. By re-propagating each sequence of diffusion-generated controls through the 6-DOF dynamics and measuring the resulting position error, we obtain a mean position error of $7.19\pm3.45\,\mathrm{m}$ for the approach cone model and $5.88\pm3.64\,\mathrm{m}$ for the LOS model. Although this dynamic consistency can improve with further model training, the discrepancy motivates the use of diffusion-generated trajectories as warm starts to traditional solvers, which can restore dynamic feasibility. While diffusion models can produce trajectories that are comparable to the SCP solutions, they offer no formal guarantee of dynamic feasibility or optimality. Instead, the diffusion model output can be used to warm-start a traditional optimal solver like the SCP solver used to generate the training data, such that the SCP solver can recover a dynamically feasible and locally optimal solution.

\begin{figure}[H]
    \centering
    \includegraphics[width=1\linewidth]{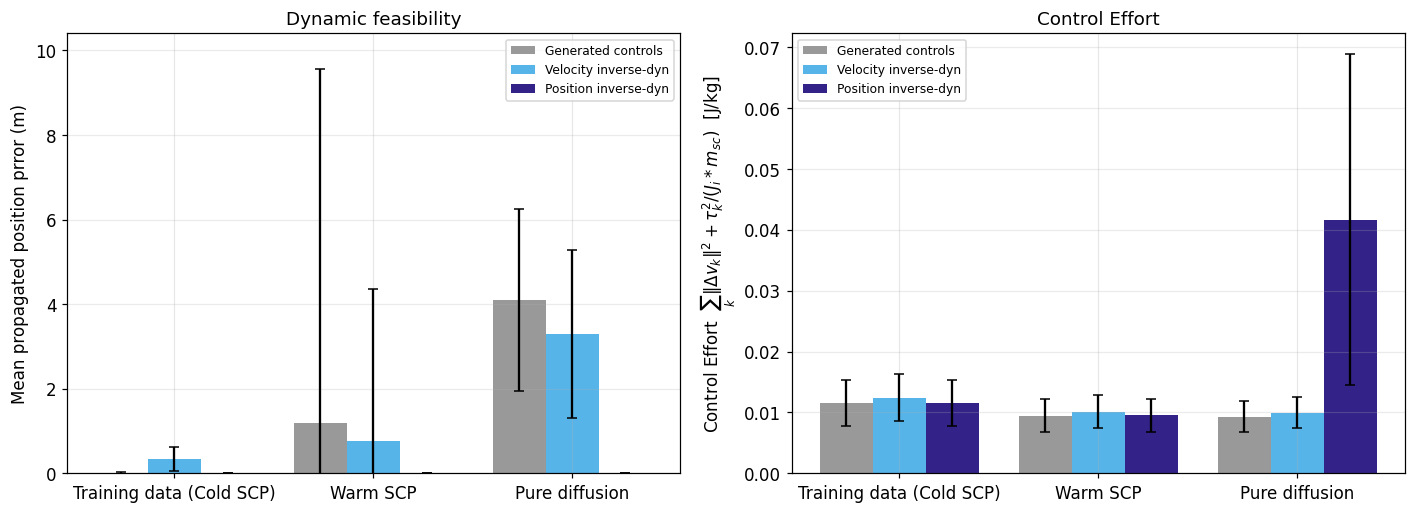}
    \caption{Dynamic Feasibility and Control Effort of Cold-Started SCP, Warm-Started SCP, and Pure Diffusion Samples Across Control Configurations (Generated, Velocity-Based Inverse Dynamics, Position-Based Inverse Dynamics)}
    \label{fig:dyn_feas}
\end{figure}

\begin{figure}[H]
    \centering
    \includegraphics[width=\linewidth]{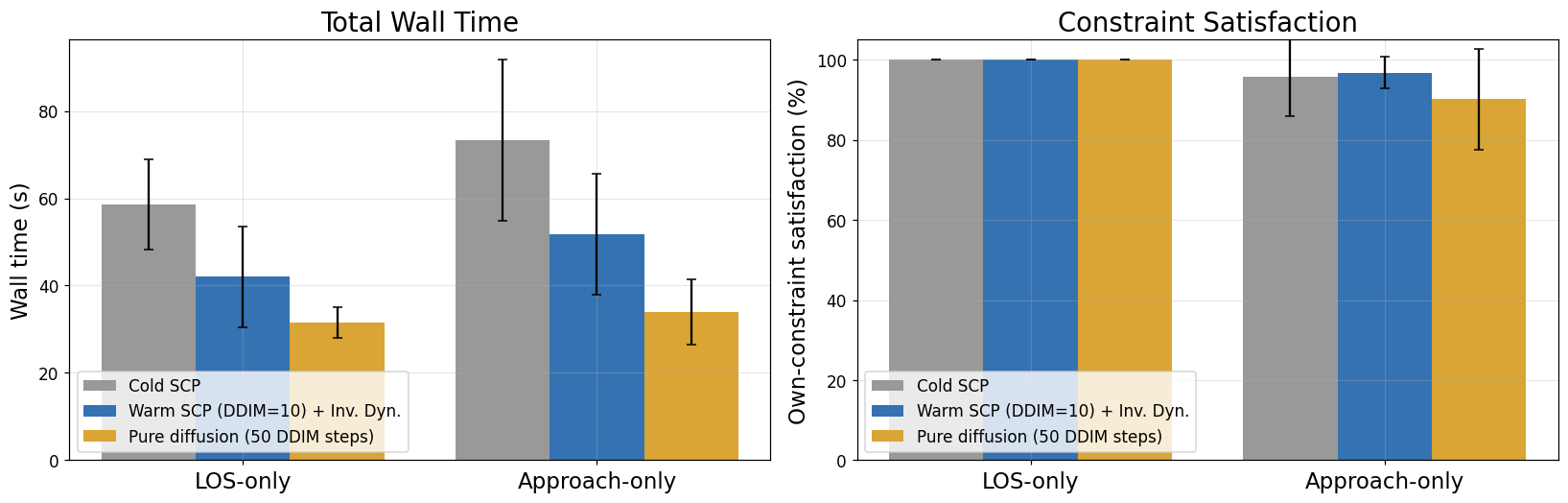}
    \caption{Total Wall Time and Constraint Satisfaction of Approach Cone and LOS Models Across Cold-Started SCP, Warm-Started SCP, and Pure Diffusion Samples}
    \label{fig:warm_time}
\end{figure}

While the warm-start framework improves mean propagated position errors on generated controls by over 70\%, the warm-start process alone does not recover dynamic feasibility that is comparable to the training data. As such, velocity-based inverse dynamics and position-based inverse dynamics are explored as alternatives to direct application of diffusion-generated control outputs. Figure \ref{fig:dyn_feas} presents the mean propagated position error and total control effort across each propagation approach, where control effort accounts for both translational and rotational actuation. We find that the combination of diffusion-warm-started SCP and subsequent controls regeneration via position-based inverse dynamics yields the best balance between error and control minimization. In Figure~\ref{fig:warm_time}, we provide a comparison of solve time, fuel optimality, and constraint satisfaction across cold-started SCP, warm-started SCP, and diffusion-only approaches. Here, the warm-started SCP solve time includes diffusion sampling time with 10 DDIM steps, SCP solve time with diffusion-output initialization, and controls computation via position-based inverse dynamics.

Over a sweep of $100$ seeds per model with $10$ denoising steps, the warm start combined with position-based inverse dynamics controls reduces the mean end-to-end solve time by $28.3\%$ for the LOS model and $29.4\%$ for the approach cone model. Furthermore, with both the cold-start and warm-start approaches, the constraint satisfaction rate remains consistent. The cold-start SCP process achieves a constraint satisfaction of $100.0\%$ for the LOS constraint and $95.9\%$ for the approach cone constraint, while the warm-start approach achieves $100.0\%$ for the LOS constraint and $96.8\%$ for the approach cone constraint. Furthermore, these time improvements are accompanied by less than a 3.5\% increase in SCP-converged control effort for the LOS model and $0.4\%$ for the approach cone model. 

The pure diffusion samples were generated with 50 DDIM steps and the performance metrics are provided as a point of comparison. While the sampling speed is much faster and the constraint satisfaction is comparable, the pure diffusion approach does not guarantee optimality or dynamic consistency. On the other hand, the warm-started SCP and position-based inverse dynamics approach achieves comparable fuel consumption and constraint satisfaction, and converges nearly 30\% faster compared to the cold-started SCP approach.

\subsection{Joint Constraint Satisfaction}

\begin{figure}[H]
    \centering
    \includegraphics[width=0.4\linewidth]{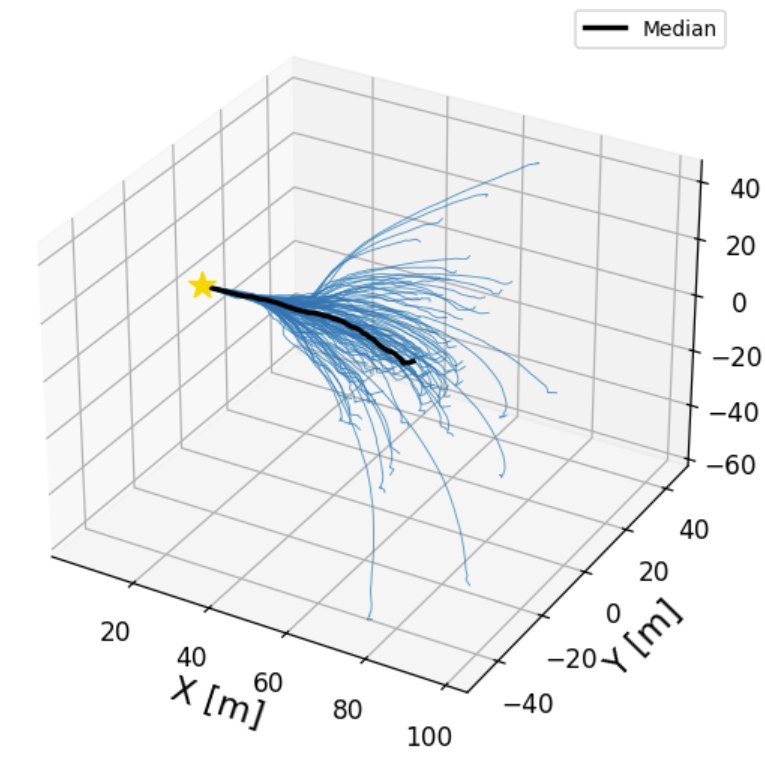}
    \caption{3D plot of Product-Composed Trajectory Samples (N=100)}
    \label{fig:prod_3d}
\end{figure}

The product-composed model was obtained according to the following energy composition:

\begin{equation}
E_{\mathrm{prod}}(x) =  0.1 \, E_{LOS,\; 6\text{-}DOF}(x) + 0.9 \, E_{appch,\; 3\text{-}DOF}(x), 
\label{eq: prod}
\end{equation}

where $E_{LOS,\;6\text{-}DOF}(x)$ denotes the sensor line-of-sight energy model and $E_{appch,\;3\text{-}DOF}(x)$ denotes the translational subspace of the 6-DOF approach cone energy model. The weights were selected based on a parameter sweep analysis and act on the shared translational dimensions while the rotational dimensions are defined solely by the LOS model. We validate empirically that the rotational policy learned by the LOS model generalizes to the approach cone-constrained trajectories. The trajectory samples from the product-composed model are presented in Figure \ref{fig:prod_3d}.

To evaluate the ability of the composed model to enforce multiple constraints simultaneously, we analyze approach cone and sensor line-of-sight (LOS) satisfaction percentage and peak violations over each trajectory’s constraint-active region. Table~\ref{tab:prod_errors} summarizes per-constraint performance for the product-composed model and combined constraint performance across all models. The product-composed model demonstrates strong multi-constraint behavior, achieving low violation across both constraints simultaneously.

\begin{table}[H]
    \centering
    \caption{Constraint Satisfaction and Peak Violation Metrics for the Product Composed Model (N=100)}
    \label{tab:prod_errors}
    \begin{tabular}{lcccc}
    \hline
     & \multicolumn{2}{c}{Approach Cone Constraint} & \multicolumn{2}{c}{Sensor LOS Constraint} \\
    Model & Satisfaction (\%) & Peak Violation ($^\circ$) & Satisfaction (\%) & Peak Violation ($^\circ$) \\
    \hline
    Product  & $99.3 \pm 5.1$ & $0.03 \pm 0.21$   & $98.6 \pm 7.1$ & $0.36 \pm 1.76$ \\
    \hline
    \end{tabular}
    
\end{table}

\begin{figure}[H]
    \centering
    \includegraphics[width=
    \linewidth]{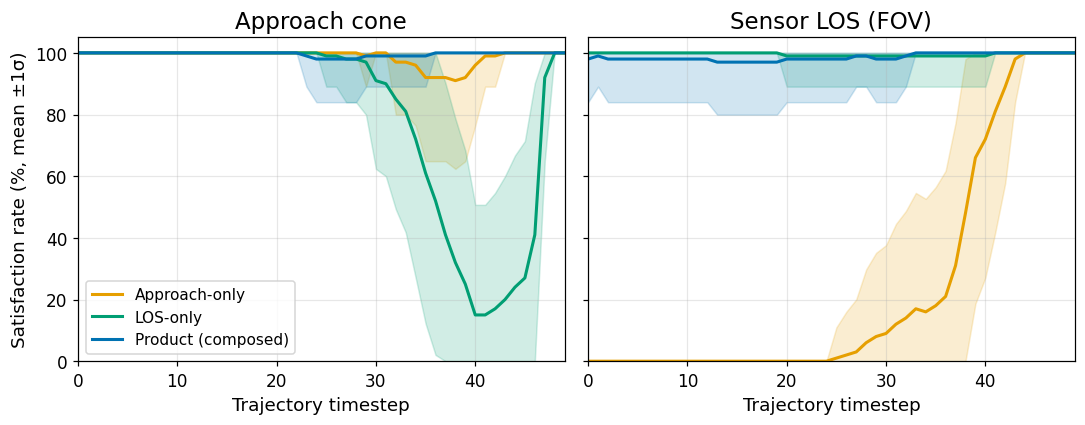}
    \caption{Comparison of Constraint Satisfaction Along Trajectory}
    \label{fig:prod satisfy}
\end{figure}

The product-composed model satisfies the approach cone constraint at $99.3\pm5.1\%$ of constraint-active timesteps and the sensor-LOS constraint at $98.6\pm7.1\%$. Each single-constraint model, in contrast, satisfies only its trained constraint: the approach cone model holds the cone at $96.1\%$ but the LOS constraint at just $24.2\%$, and the LOS model holds LOS constraint at $99.6\%$ but the approach cone constraint at only $36.9\%$, as presented in Table \ref{tab:constraint_errors}. The composed model thus attains specialist-level satisfaction on both constraints from a single sampler, matching or exceeding the approach cone specialist on the cone ($99.3\%$ vs.\ $96.1\%$) while trailing the LOS specialist by roughly one point ($98.6\%$ vs.\ $99.6\%$). The constraint satisfaction rate over the course of the trajectory is visualized in Figure \ref{fig:prod satisfy}

Considering the joint satisfaction of the approach cone and sensor LOS constraints, we see that the product-composed model demonstrates strong multi-constraint behavior, achieving low violation across both constraints simultaneously. While each single model achieves minimal error on its respective trained constraint, both exhibit substantial violations when evaluated jointly. These results indicate that product composition can effectively combine probabilistic solution regions to encourage joint constraint feasibility. The composed model matches or exceeds each specialist on the approach cone constraint while remaining comparable to the dedicated LOS model within 1 percentage point, yielding a balanced solution that satisfies both constraints in the evaluated samples.

\begin{figure}[H]
    \centering
    \includegraphics[width=1\linewidth]{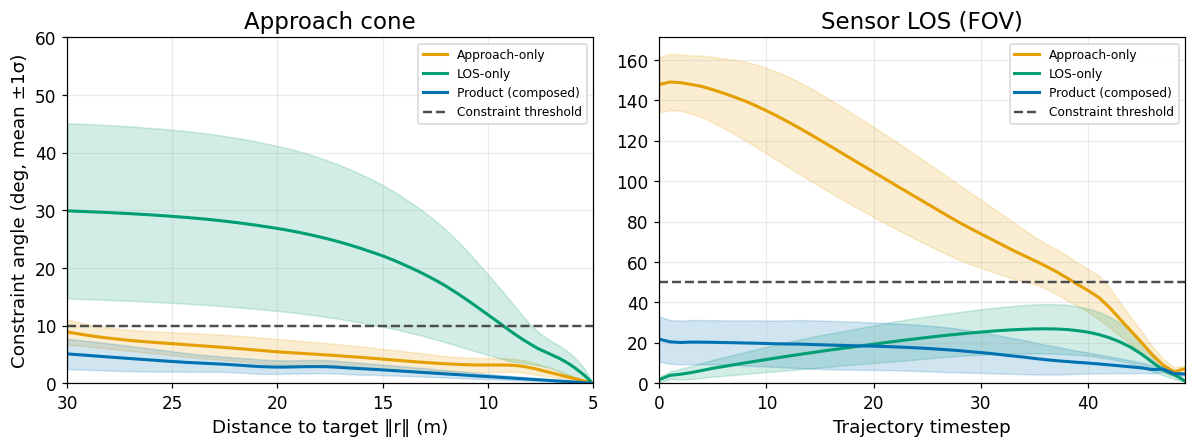}
    \caption{Approach Angle and Sensor Line-of-Sight Angle Over Constraint-Active Trajectory Region}
    \label{fig:prod angle}
\end{figure}

Figure \ref{fig:prod angle} illustrates how the constraint metrics evolve over the constraint-active portion of the trajectory. The product-composed model maintains approach angles below the $10^\circ$ threshold for the majority of the trajectory and exhibits a smooth decrease as the vehicle approaches the target. In contrast, the sensor LOS model shows sustained violations of the approach constraint, and the approach-only model shows mixed satisfaction of the approach angle constraint. As for the LOS angle evolution, the product-composed model maintains pointing angles well within the $50^\circ$ field-of-view constraint throughout the trajectory, while significantly reducing the large initial pointing errors observed in the approach-only model. Although the dedicated LOS model achieves the lowest absolute error, the composed model demonstrates comparable constraint satisfaction with improved balance across both constraints.

Taken together, these results indicate that the composed model successfully considers both constraints simultaneously, achieving low mean error and improved constraint performance over the full trajectory, without being explicitly trained on the joint constraint space.

\subsection{Synthetic Constraint Composition}
To demonstrate the flexibility of the framework, we compose the approach cone energy model with a synthetic obstacle avoidance energy landscape. Rather than training on obstacle avoidance scenarios, we inject an analytically-derived energy field at inference time, demonstrating how the framework can incorporate new constraints without retraining. The composed energy function is
\begin{equation}
E_{\text{Neg}} = E_{\text{appch}} - 0.5 E_{\text{obs}},
\end{equation}
where $E_{\text{obs}}$ is defined in Equation \ref{eq:obs-energy} and parameters are specified in Table \ref{tab:obs_params}. 

\begin{table}[H]
    \centering
    \caption{Obstacle Constraint Parameters}
    \label{tab:obs_params}
    \begin{tabular}{ll}
    \hline
    Parameter & Value \\
    \hline
    Obstacle Center $c$ (m) & (80, 0, 0)\\ 
    Obstacle Radius $R$ (m) & 10 \\
    Softening Length $\sigma$ (m) & 10 \\
    Gain $g$ & 0.5 \\
    \hline
    \end{tabular}
\end{table}

\begin{figure}[H]
    \centering
    \includegraphics[width=\linewidth]{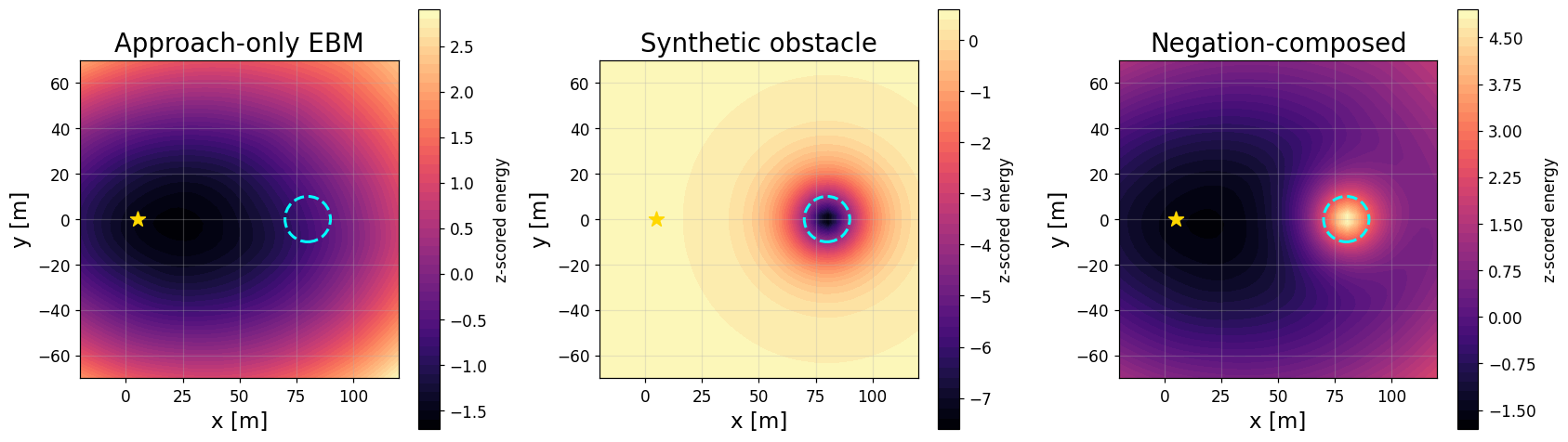}
    \caption{Energy Landscape ($z=0$, Z-score normalized for visualization) for the Approach Cone Model, Synthetic Obstacle Model, and Negation Composition}
    \label{fig:obs_energy}
\end{figure}

Figure~\ref{fig:obs_energy} visualizes the approach cone energy, the synthetic obstacle barrier, and their negation composition on an $xy$ grid at $z=0$. The learned approach cone energy defines a natural corridor through which trajectories should pass, while the synthetic obstacle term provides a heuristic guidance mechanism that deflects samples away from the high-energy collision region.

\begin{figure}[H]
    \centering
    \includegraphics[width=\linewidth]{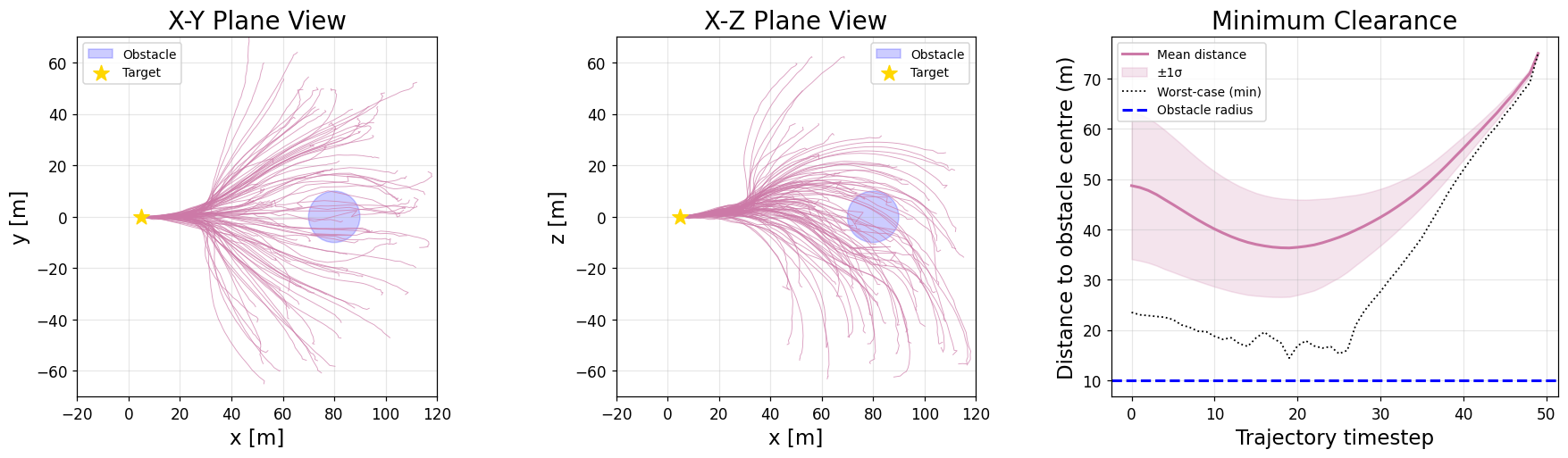}
    \caption{Plane view of Negation-Composed Trajectories and Mean Obstacle Distance Over Trajectory Time}
    \label{fig:obs_satisfy}
\end{figure}

The composed model achieves strong constraint satisfaction across 100 samples: 100\% obstacle avoidance with mean minimum clearance of 22.9m ($\pm 9.6$m), while maintaining 96.5\% approach cone satisfaction. Figure~\ref{fig:obs_satisfy} shows an $xy$-plane and $xz$-plane view of the trajectories and the mean obstacle distance over time. The samples curve around the obstacle and generally hold a generous margin, as encouraged by the gradual taper of the obstacle energy. While an aggressive synthetic barrier can introduce localized jaggedness in the trajectory, this can be remedied by further tuning of the obstacle energy parameters or by adopting the warm-start approach.

\begin{table}[H]
    \centering
    \caption{Constraint Satisfaction, Peak Violation, and Minimum Clearance Metrics for the Obstacle-Avoidance Composed Model (N=100)}
    \label{tab:obs_errors}
    \begin{tabular}{lcccc}
    \hline
     & \multicolumn{2}{c}{Approach Cone Constraint} & \multicolumn{2}{c}{Obstacle Avoidance Constraint} \\
    Model & Satisfaction (\%) & Peak Violation ($^\circ$) & Satisfaction (\%) & Min Clearance (m) \\
    \hline
    Obstacle & $96.5 \pm 9.5$ & $0.21 \pm 0.61$   & $100.0 \pm 0.0$ & $22.86 \pm 9.58$ \\
    \hline
    \end{tabular}
    
\end{table}

This result is significant because the obstacle constraint was never seen during training. The closed-form energy landscape is injected at sampling time and steers the learned distribution into a new feasible region, demonstrating that the diffusion-based trajectory generation framework extends beyond composing pre-trained models to composing learned and hand-specified constraints.

\section{Discussion}
This framework couples diffusion-based generative modeling with inference-time energy composition to produce constraint-modular RPO trajectories. The presented results support three claims.

First, diffusion models can generate constraint-satisfying RPO trajectories that reduce end-to-end solution time when used to warm-start a traditional SCP solver. The diffusion-warm-started solve with position-based inverse dynamics reduces end-to-end wall-clock time by roughly $30\%$ (including the cost of generating the diffusion sample) at a control-effort penalty of $\le3.5\%$ and with constraint satisfaction preserved. Notably, this speed-up stems less from a reduction in SCP iterations than from better-conditioned convex subproblems. The constraint-aware initial guess places the SCP solver closer to the feasible manifold and thus lowers the per-iteration cost. Furthermore, the near-identical converged fuel cost confirms that the warm start trades little to no optimality for speed. In short, the diffusion sample supplies a constraint-aware, near-optimal initial guess that can be paired with a traditional solver to certify dynamic feasibility and local fuel optimality.

Second, composition can induce combined constraint behavior without retraining and with minimal constraint penalty. Product composition yields a model that satisfies the approach cone and sensor-LOS constraints at $99.3\%$ and $98.6\%$ of constraint-active timesteps respectively, despite each single-constraint model satisfying the competing constraint at only $24$–$37\%$. Relative to each constraint-specific model, the product modestly exceeds the approach cone constraint satisfaction by 3.2 percentage points, while incurring a slight reduction of 1.0 percentage points in the satisfaction of the LOS constraint. 

Third, learned energy landscapes can be composed with synthetic energies to extend the framework beyond pre-trained models. Across $100$ samples, the negation-composed model achieves $100\%$ satisfaction of the obstacle-avoidance constraint while preserving the base approach cone satisfaction at $96.5\%$ of constraint-active timesteps. Because the obstacle term is analytic, new keep-out constraints can be specified and enforced at sampling time without obstacle-specific training data or retraining, which can be an advantageous capability for mission scenarios that demand quick adaptation to unseen obstacle configurations. However, an introduced synthetic energy can degrade trajectory smoothness depending on the barrier form and composition parameters, motivating well-tuned barriers or use as a warm start.

Several limitations qualify these findings. First, the analyses are empirical, drawn from $100$ samples on a single rendezvous scenario, and should be read as observations rather than guarantees. The warm-start approach remains the only component that certifies dynamic feasibility and local optimality. Additionally, the composition weights and energy parameters were manually tuned for this work, and a real deployment would require systematic calibration. In particular, when constraints act on different state subsets or bind over overlapping time windows, naive weighting can let some energy fields dominate and suppress joint satisfaction. Finally, dynamic consistency is enforced only weakly during generation and is not explicitly considered during composition. The product-composed samples exhibit larger re-propagation error than the single-constraint samples, and should therefore be validated through the warm-start optimizer.

Future work could include physics-informed training and composition to tighten dynamic feasibility and fuel optimality without the warm-start framework. Additionally, the synthetic energy approach can be expanded beyond obstacle avoidance, to constraints like keep-out and corridor constraints. Finally, further study of the composition method could explore principled weight calibration or additional composition operators for new constraint types.

\section{Conclusion}
In this paper, we present a diffusion-based RPO trajectory generation approach, in which independently trained, single-constraint energy-based models are composed at sampling time to enforce unseen constraint combinations without additional training. Three findings support this framework. First, the single-constraint models can reproduce intended constraint behavior with high fidelity. Second, product composition of the learned models can yield joint satisfaction of nontrivial constraints. Third, composing a learned energy with a closed-form energy field can extend the framework to unseen constraints.

Since diffusion samples carry no formal feasibility or optimality guarantee, we pair them with the same sequential convex programming solver used to generate the training data, warm-starting it to cut end-to-end wall-clock time by roughly $30\%$ with minimal control-effort penalty and with constraint satisfaction preserved, yielding a dynamically feasible, locally fuel-optimal maneuver. Inference-time composition thus provides a flexible, retraining-free mechanism for multi-constraint RPO trajectory generation, and the diffusion-warm-started SCP pipeline provides a practical route from constraint-aware samples to deployable trajectories. Broader initial-condition and geometry sweeps, systematic weight calibration, and dynamics-informed training and composition remain as future work.

\section{Acknowledgments}
Research was sponsored by the Department of the Air Force Artificial Intelligence Accelerator and was accomplished under Cooperative Agreement Number FA8750-19-2-1000. The views and conclusions contained in this document are those of the authors and should not be interpreted as representing the official policies, either expressed or implied, of the Department of the Air Force or the U.S. Government. The U.S. Government is authorized to reproduce and distribute reprints for Government purposes notwithstanding any copyright notation herein.

\appendix

\bibliographystyle{AAS_publication}   
\bibliography{references}   

\end{document}